\documentclass[10pt,twocolumn,letterpaper]{article}

\usepackage{iccv}
\usepackage{times}
\usepackage{epsfig}
\usepackage{graphicx}
\usepackage{amsmath}
\usepackage{amssymb}
\usepackage{subcaption}
\usepackage{gensymb}
\usepackage[dvipsnames]{xcolor}

\usepackage[breaklinks=true,bookmarks=false]{hyperref}

\iccvfinalcopy
\def\iccvPaperID{5134} 

\begin{document}

\title{Floors are Flat: Leveraging Semantics for Real-Time Surface Normal Prediction}

\author{Steven Hickson\\
Google, Georgia Tech\\
{\tt\small shickson@gatech.edu}
\and
Karthik Raveendran\\
Google\\
{\tt\small krav@google.com}
\and
Alireza Fathi\\
Google\\
{\tt\small alirezafathi@google.com}
\and
Kevin Murhpy\\
Google\\
{\tt\small kpmurphy@google.com}
\and
Irfan Essa\\
Google, Georgia Tech\\
{\tt\small irfan@cc.gatech.edu}
}

\newcommand{\eat}[1]{}

\maketitle

\begin{abstract}
    We propose 4 insights that help to significantly improve the performance of deep learning models that predict surface normals and semantic labels from a single RGB image. These insights are: (1) denoise the "ground truth" surface normals in the training set to ensure consistency with the semantic labels; (2) concurrently train on a mix of real and synthetic data, instead of pretraining on synthetic and finetuning on real; (3) jointly predict normals and semantics using a shared model, but only backpropagate errors on pixels that have valid training labels; (4) slim down the model and use grayscale instead of color inputs. Despite the simplicity of these steps, we demonstrate consistently improved results on several datasets, using a model that runs at 12 fps on a standard mobile phone.
    \eat{
    
    We introduce 4 simple tricks to significantly improve the performance of deep learning models that predict surface normals and semantic labels from a single RGB image. The tricks are: (1) denoise the "ground truth" surface normals in the training set by ensuring consistency with the semantic lables; (2) concurrently training on a mix of real and synthetic data, instead of pretraining on synthetic and finetuning on real; (3) jointly predicting normals and semantics using a shared model, but only backpropagating errors on pixels that have valid training labels; (4) slimming down the model and using gray scale instead of color inputs.
    Despite the simplicity of these methods, we demonstrate
    consistently improved results on several datasets,
    using a model that runs at 12 fps on a standard mobile phone.

    significant improvements in accuracy compared to the state of the art method of Qi et al \cite{qi2018geonet}. In particular, on the NYU test set, we reduce the mean angle error from 19 to 17 degrees, and increase the accuracy  from 79.5\% to 81.2\%.
    Importantly, our method does this while running at 12 fps on a standard mobile phone.
    }
\end{abstract}

\eat{
\begin{abstract}
   We introduce a method that showcases a significant improvement in performance of surface normal prediction from a single RGB image. Our method relies on training over a mixture of real and synthetic data that is corrected leveraging semantics of the scene, which makes the data much more closely match realistic surface normals. Furthermore, we show that joint prediction of semantic labels and surface normals concurrently improves both labeling and normals prediction. We use these insights to create an efficient and accurate model that is capable of computing robust surface normals at 12 fps on a current generation mobile phone for a variety of real-time vision applications.
\end{abstract}
}

\eat{

\begin{abstract}
   We show that 
   we can significantly improve the performance of standard methods for surface normal prediction
   from RGB images by training on a mixture of real and synthetic data that we correct using semantics so that it more closely matches realistic surface normals. Furthermore, we show that jointly predicting semantic labels and surface normals  concurrently improves both results. We then use these insights to create an efficient and accurate model that is capable of computing robust surface normals at 12 fps on a mobile phone for deep learning applications.
\end{abstract}

\begin{abstract}
   In this paper, we propose a way to achieve robust surface normal estimation from a single RGB image on a mobile device. We demonstrate that current surface normal estimation methods train on substandard and noisy ground truth data resulting in poor performance. Results can be improved with ground truth training data that more closely matches realistic surface normals. We also show the effects of training on multiple datasets mixed together, including mostly synthetic data to reduce dataset bias and create smoother, more realistic surface normals. Further, we show that a joint prediction task with semantic labeling and surface normals learned concurrently improves both results. The insights above can be combined to create an efficient and accurate model that is capable of computing robust surface normals at 12 fps on a mobile phone.
\end{abstract}
}
\section{Introduction}

In this paper, we address the problem of learning a model that can predict surface normals and semantic labels for each pixel, given a single monocular RGB image. 
This has many practical applications, such as in augmented reality and robotics.


Most high-performing methods train deep neural networks to perform the task of estimating surface normals using large training sets.
However, an often overlooked aspect of such approaches is the quality of the data that is used for training (and testing).
We have found that the standard technique for estimating surface normals from noisy depth data, such as the widely used method
of Ladicky et al. 
\cite{zeisl2014discriminatively},
can result in inconsistent estimates for the normals of neighboring points (see Figure~\ref{fig:ladicky_error} for an example).

\begin{figure}[t]
\centering
\begin{subfigure}{0.22\textwidth}
	\includegraphics[width=\textwidth]{fig/rgbImage_255.png}
    \label{fig:rgb_image}
\end{subfigure}
\begin{subfigure}{0.22\textwidth}
	\includegraphics[width=\textwidth]{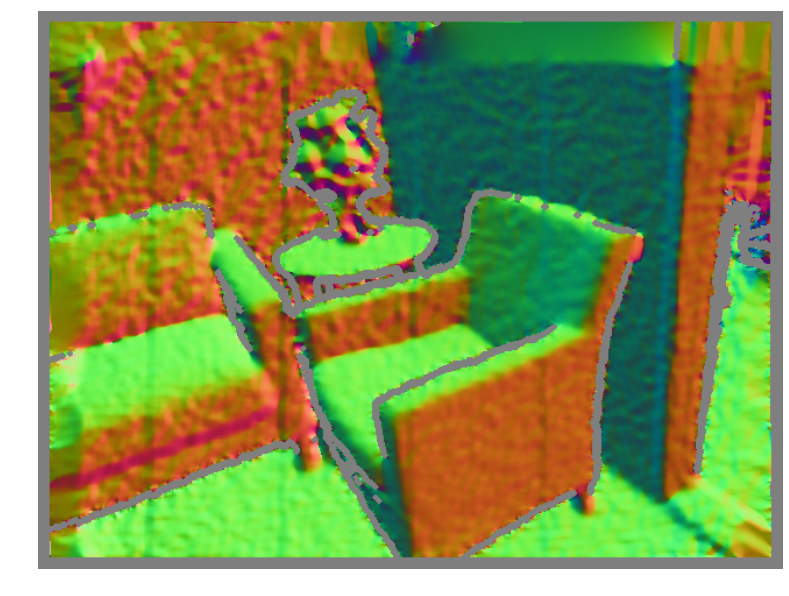}
    \label{fig:unsmoothed_normals}
\end{subfigure}
\begin{subfigure}{0.22\textwidth}
	\includegraphics[width=\textwidth]{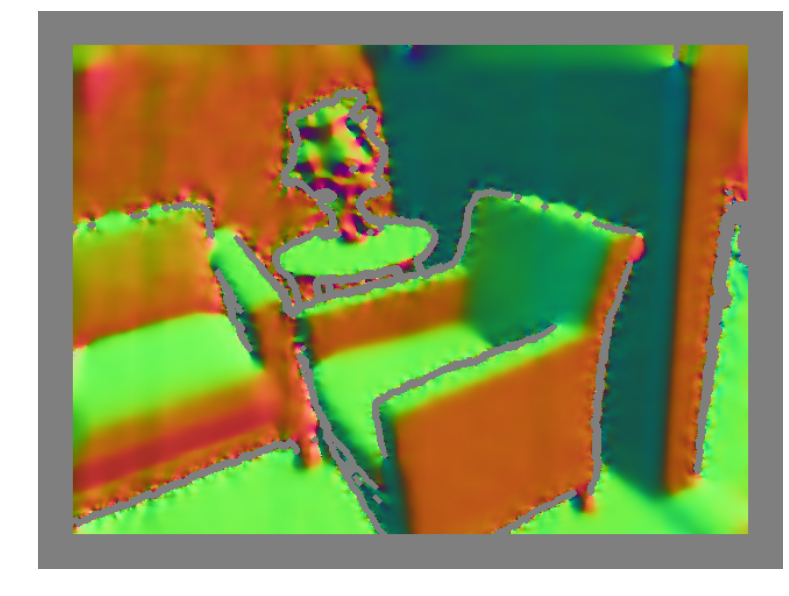}
    \label{fig:smoothed_normals}
\end{subfigure}
\begin{subfigure}{0.22\textwidth}
	\includegraphics[width=\textwidth]{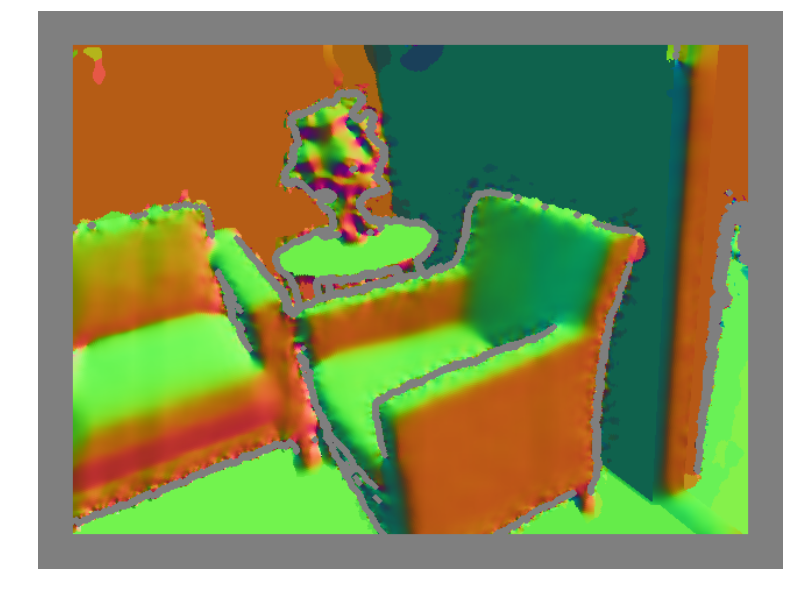}
    \label{fig:semantic_smoothed_normals}
\end{subfigure}
\caption{Visualization of different ways of
computing "ground truth" normals.
Top left:  a sample image from the NYUDv2 dataset.
Top-right: computed using method similar to \cite{Eigen_2015_ICCV} with a small window.
Bottom-left: results of our  method using larger depth-adaptive smoothing.
Bottom-right: results of our method
after semantic smoothing (if labels are available). Note that the back and right wall are cleaned up to a large degree due to this correction.}
\label{fig:teaser}
\end{figure}

We propose a simple technique to fix this,  by regularizing the prediction of normals that correspond to the same surface.
This encodes our intuition that floors should be flat and pointing up, etc.
To estimate which pixels belong to the same surface, we leverage the fact that many depth datasets also have per-pixel semantic labels.
This in itself does not tell us which facet of the object a pixel belongs to, but we use simple heuristics to solve this problem, as described in Section~\ref{sec:normal_computation}.
See Figure~\ref{fig:teaser} for an illustration of the benefits of this approach.

Unfortunately, even after such ``data cleaning'',
most real world datasets are still too small to train deep models, so it has become popular to leverage synthetically generated images. These are noise-free, but it is not obvious how best to combine real and synthetic data.
The standard practice (e.g., \cite{mccormac2017scenenet}) is to pretrain on the synthetic Scenenet dataset and then fine tune on the real dataset. 
We propose a simple improvement to this idea, which is to train the model on a carefully chosen mix of real and synthetic images in each minibatch. This simple insight improves results considerably, as we show in Section~\ref{sec:mixing_data}.

In addition to improving the way data is used, we propose some improvements to standard modeling methods.
In particular, we train a model
to  jointly predict surface normals and semantic labels, using an encoder-decoder network with two output heads.
We take care to only backpropagate errors on outputs for which we have labels.
Although this idea of multi-task training is not new,
we are the first to show (as far as we know)
a benefit from this approach as measured by improved performance on both tasks, as we discuss in  Section~\ref{sec:joint}.

Finally, since most of the applications of depth estimation require a real-time method, we describe some simple tricks to make our model much smaller and faster, with negligible loss in accuracy.
The result is a method that can run at 12fps on a standard mobile phone, while achieving state-of-the-art accuracy on standard benchmarks.

In summary, our main contributions are as follows:
\begin{itemize}
\itemsep0em 
  \item A simple method for computing reliable ground truth normals using "semantic smoothing".
  (We will release the cleaned-up data when the paper is accepted.)
  
  \item A simple method for training with synthetic and real data which gives state of the art results.
  
  \item A simple method that jointly learns semantics and surface normals in an end-to-end manner, increasing the accuracy of both.

  \item A simple method for making a model run in real-time (12 fps) on a mobile phone while still giving good accuracy. 
\end{itemize}
While each of these contributions in themselves is small, and arguably not new, we believe these steps deserve to be more widely known, since they are very effective to address a complex task.
 
\section{Related Work}
\label{sec:related}

Traditional methods for estimating surface normals were largely limited by sources for ground truth data, and instead incorporated explicit priors such as shading, vanishing points \cite{Hoiem2007}, or world constraints \cite{schwing_2003_ICCV}. With the advent of widely available and inexpensive depth sensors, data-driven approaches to this problem became more popular. Ladicky et al \cite{zeisl2014discriminatively} introduced a discriminatively trained learning based algorithm by combining pixel and segment-level cues. Fouhey \cite{Fouhey_2013_ICCV} and colleagues explored the use of learned sparse 3D geometric primitives and higher level constraints to predict surface normals. 

In recent years, convolutional neural networks (CNNs) have proven to be a very effective means of tackling a wide range of image-level tasks. Wang et al. \cite{wang2015designing} introduced the first CNN-based method to solve the problem of dense surface normal estimation by fusing both scene level and patch level predictions. Eigen and colleagues \cite{Eigen_2015_ICCV} were the first to predict depth, surface normals, and semantic segmentation using the same multi-scale network architecture for each task (though not jointly). Bansal et al. \cite{DBLP:journals/corr/BansalRG16} improved upon this architecture using skip connections and used it as input to jointly predict pose and style of 3D CAD models from an RGB image. In SURGE, Wang et al. \cite{wang2016surge} uses a dense CRF to refine the output of their CNN model and demonstrate higher quality results on planar regions. 

Researchers have also begun to explore the connections between various pixel level labelling tasks. Dharmasiri et al. \cite{dharmasiri2017joint} demonstrate that by having a single network jointly predict surface normals, depth, and curvature, they are able to almost match or improve upon networks tuned for these tasks independently. Similarly, Nekrasov \cite{nekrasov2018real} and colleagues explored the connections between depth estimation and semantic segmentation with a focus on the effects of asymmetric dataset sizes. Xu et al. \cite{DBLP:journals/corr/abs-1805-04409} developed a prediction and distillation network that uses multiple intermediate representations such as contours and surface normals to achieve the final task of depth estimation and scene parsing.  Similarly, Kokkinos \cite{kokkinos2017ubernet} showed that a single unified architecture is capable of learning a wide range of image labeling tasks. A couple of works \cite{qi2018geonet, yang2017unsupervised} enforce consistency between joint predictions of depth and normals. In this paper, we show that semantic segmentation can improve normal prediction when predicting both jointly which results in even higher performance on planar regions. Some work, such as \cite{liu2018planercnn} have had success purely predicting planar regions instead of surface normals, though this limits the scope of the problem.

Another promising avenue for gathering data for surface normal estimation is synthetic rendering. Zhang et al. \cite{zhang2017physically} train their normal prediction network on a large dataset of rendered images and fine-tune it on real data. Ren et al. \cite{ren2018cross} use an unsupervised domain adaptation method based on adversarial learning to transfer learned features from synthetic to real images. In this paper, we use synthetic data in our training but batch-wise mix it with real data in an end to end training setup.

\section{Computing better ground truth normals}
\label{sec:normal_computation}

In this section, we discuss two simple methods for computing better ground truth normals from (real) depth datasets.

\subsection{Datasets}

\noindent We use two commonly used real-world depth datasets.

\begin{figure*}
	\includegraphics[width=\textwidth]{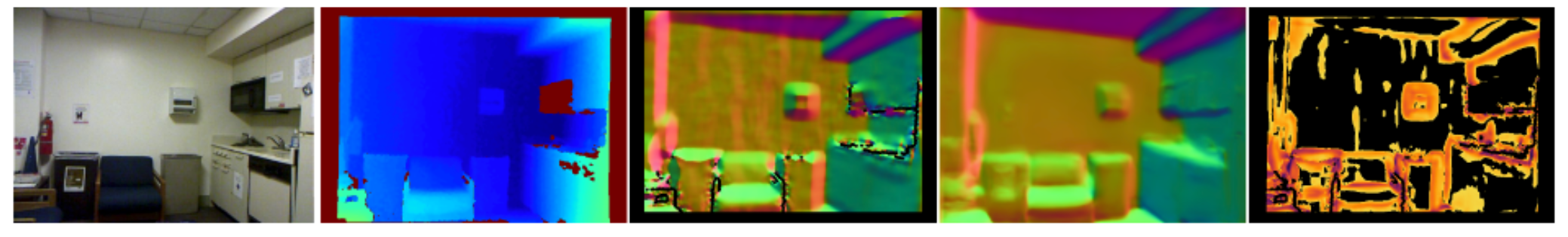}
\caption{
Illustration of the NYU data, together with the predictions of our model on them.
Columns, from left to right: RGB image, depth, ground truth surface normals, our predictions, error image (where black is under 11.25 degrees errors, and then error increases from yellow to purple).
}
\label{fig:nyu}
\end{figure*}

\begin{figure*}
	\includegraphics[width=\textwidth]{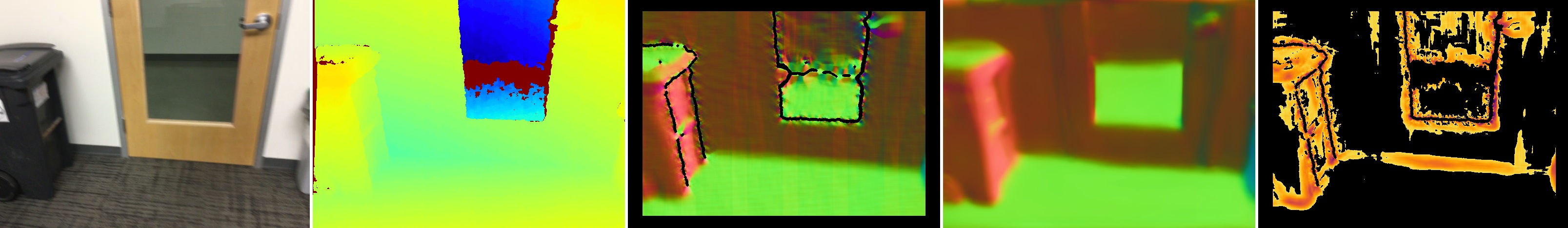}
\caption{
Qualitative evaluation on the Scannet data. Columns are the same as in Figure~\ref{fig:nyu}.
}
\label{fig:scannet}
\end{figure*}

\noindent\textbf{NYUDv2}~\cite{Silberman:ECCV12} is a dataset of indoor environments taken with a Kinect device which results in approximately 450,000 640x480 RGB-depth pairs. 1449 of these images, split into a predefined train/test, have the depth pre-processed and have semantic labels for each pixel. This is the dataset used by most methods for evaluation.
See Figure~\ref{fig:nyu} for an example.


\noindent\textbf{Scannet}~\cite{dai2017scannet} is a dataset of approximately 2 million 640x480 RGBD sensor images that also include pixel-level semantic segmentation. Instead of 2D annotation like in NYUDv2, these are done in the full 3D environment and projected back into 2D. This allows it to have many more annotations compared to NYUDv2; however, due to this method, the annotations and edges don't match up as perfectly.
See Figure~\ref{fig:scannet} for an example.


\subsection{Problems with current techniques}

\begin{figure}[b]
\centering
 \begin{subfigure}{0.22\textwidth}
	\includegraphics[width=\textwidth]{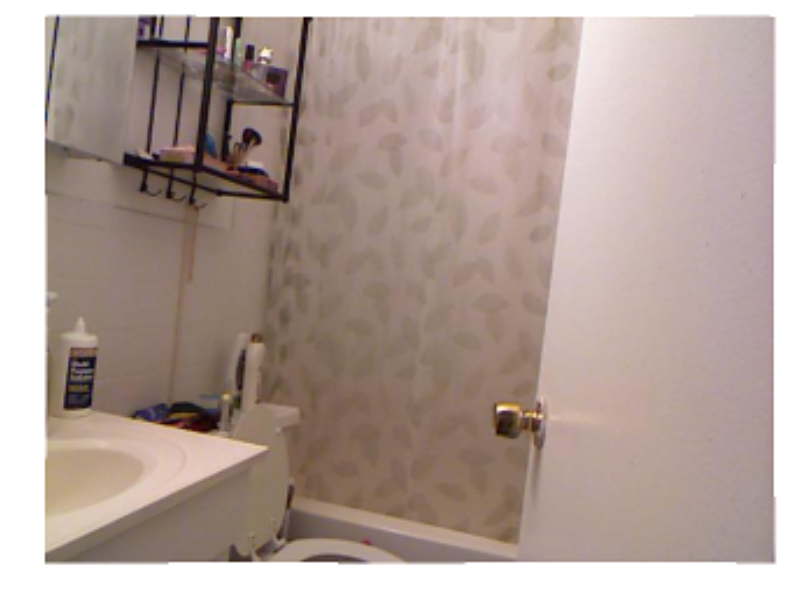}
\end{subfigure}
\begin{subfigure}{0.22\textwidth}
	\includegraphics[width=\textwidth]{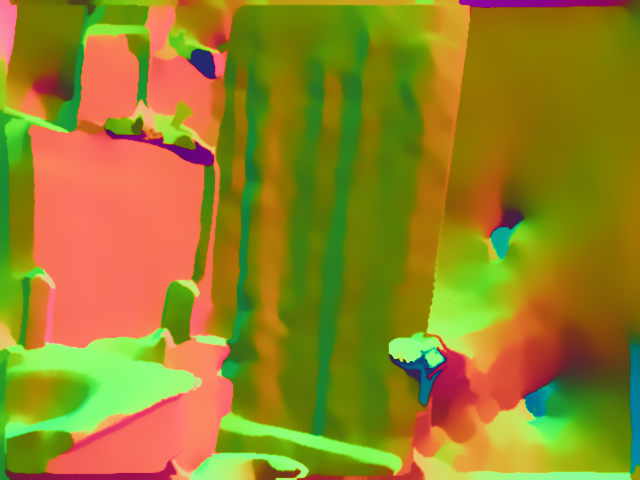}
\end{subfigure}
\caption{A common example of the errors seen in the normals from \cite{zeisl2014discriminatively} that many use for training and evaluation. In these visualizations, (r, g, b) map to (x, y, z) of the normal at that location. Note the oversmoothing, which reduces and removes the normals of small objects. This image also demonstrates why it is important to only backpropagate on pixels that have valid depth, as the right side of the image has incorrect normal data due to noisy and missing depth values.}
\label{fig:ladicky_error}
\end{figure}

Surface normals for real world datasets are typically derived from the depth data captured by commodity depth sensors or stereo matching algorithms. For instance, the NYUDv2 dataset was captured using a Kinect v1, while ScanNet uses a similar Structure sensor. These sensors are well known to suffer from axial noise which is related to the distance of the surface from the sensor. As a consequence, surface normals that are computed from this data tend to exhibit artifacts that are noticeable,
especially on distant planar regions.

\begin{figure}[t]
\centering
\begin{subfigure}{0.22\textwidth}
	\includegraphics[width=\textwidth]{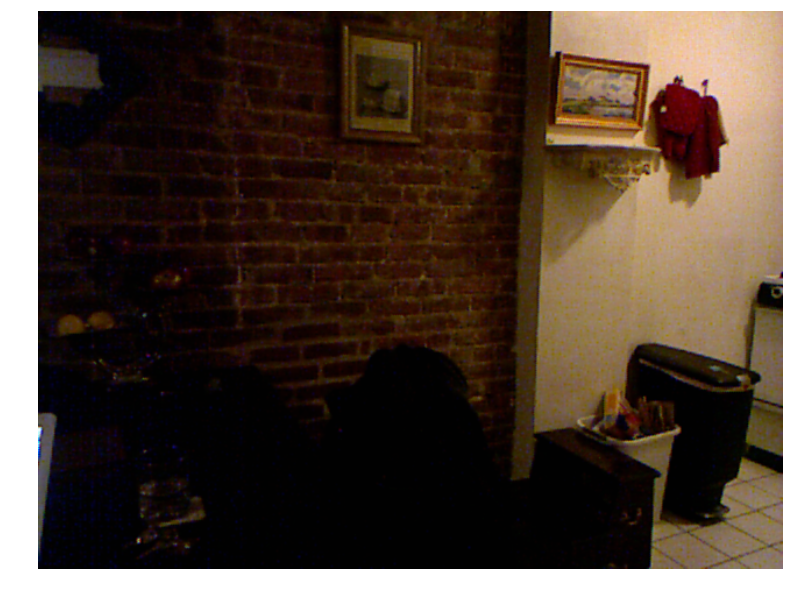}
	\caption{ }
    \label{fig:normals_rgb}
\end{subfigure}
\begin{subfigure}{0.22\textwidth}
	\includegraphics[width=\textwidth]{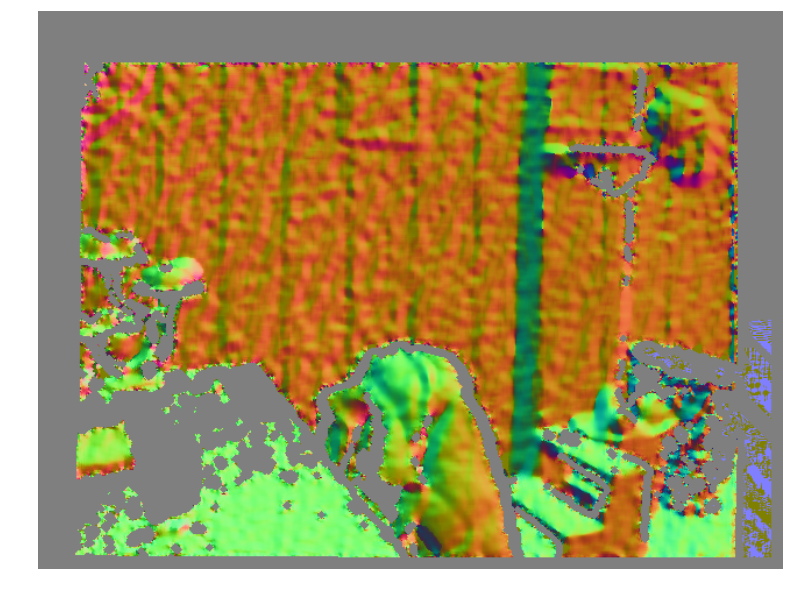}
	\caption{ }
    \label{fig:normals_raw}
\end{subfigure}
\begin{subfigure}{0.22\textwidth}
	\includegraphics[width=\textwidth]{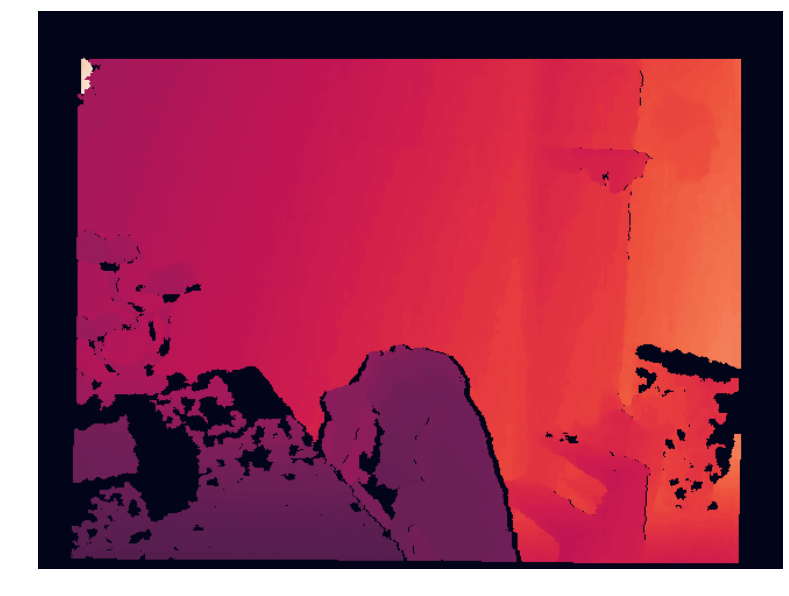}
	\caption{}
    \label{fig:normals_depth}
\end{subfigure}
\begin{subfigure}{0.22\textwidth}
	\includegraphics[width=\textwidth]{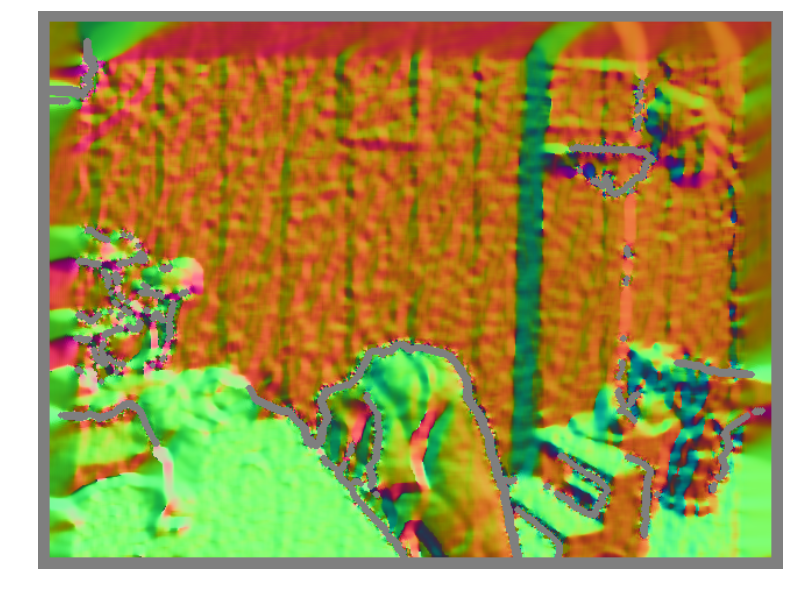}
	\caption{}
    \label{fig:normals_small}
\end{subfigure}
\begin{subfigure}{0.22\textwidth}
	\includegraphics[width=\textwidth]{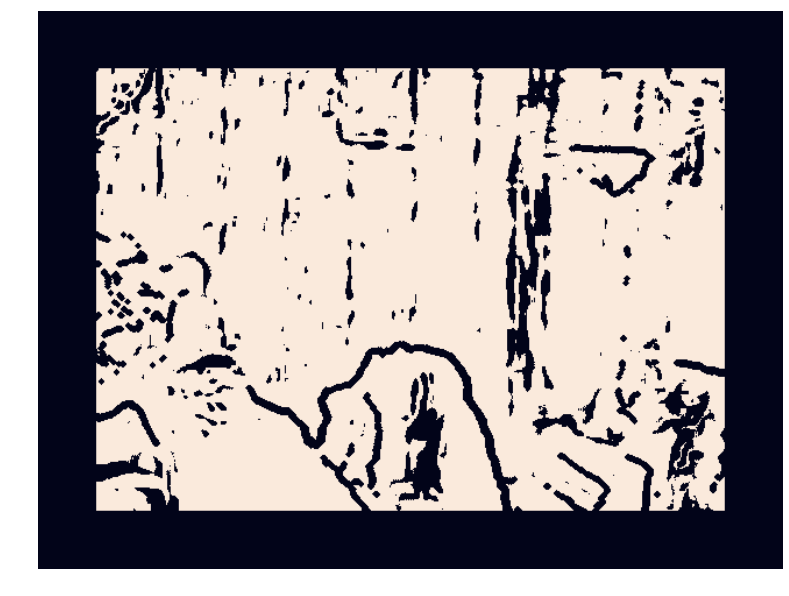}
	\caption{}
    \label{fig:normals_mask}
\end{subfigure}
\begin{subfigure}{0.22\textwidth}
	\includegraphics[width=\textwidth]{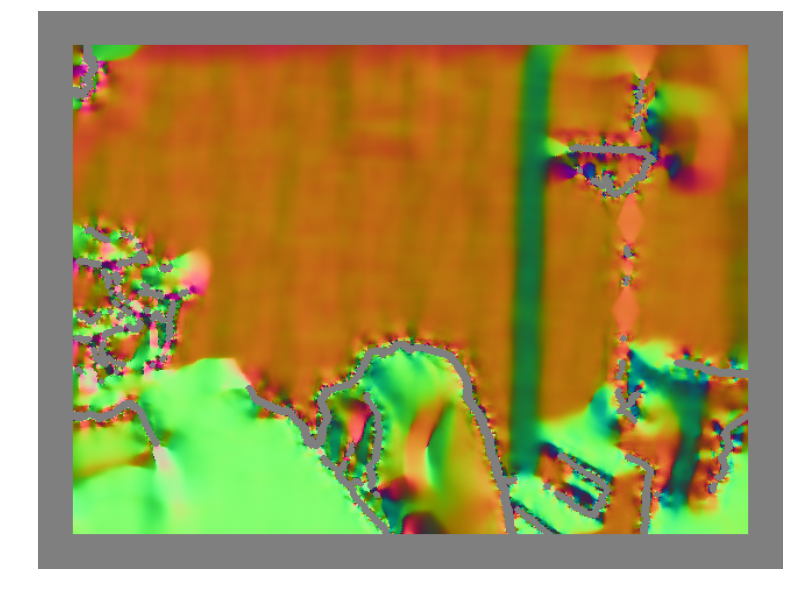}
	\caption{}
    \label{fig:normals_big}
\end{subfigure}
\caption{Effect of window smoothing size on surface normals. a) RGB image. b) Normals computed with a normal smoothing size of 10. c) In-painted depth. d) Normals computed similar to \cite{Silberman:ECCV12} from (c) using window size 10. e) The mask generated for valid training pixels. f) The normals with window size 30 that we train on (backpropagating only when e is valid).}
\label{fig:normals_error}
\end{figure}

Broadly speaking, prior work has used one of two normal estimation methods to generate ground truth for training: 
least-squares estimation on a per-point basis using RANSAC after de-noising the point cloud \cite{zeisl2014discriminatively},
or
local plane computation using the covariance matrix over a window \cite{Silberman:ECCV12}.
In our experiments, we have found both of these approaches result in ground truth normals that have considerably more errors. For instance, in
Figure~\ref{fig:ladicky_error}, the former method produces oversmoothed and incorrectly oriented planar patches on regions like the sink,
while in 
Figure~\ref{fig:normals_error}(d), the latter method produces highly noisy planar surfaces. We hypothesize that this could cause inferior results when used to train. In Table~\ref{table:different_gt}, we show that training on noisy ground truth normals results in a noticeable drop in accuracy. Training on data using the smaller smoothing window of ~\cite{Silberman:ECCV12} results in a mean angle error of 27.5 degrees, compared to 22 degrees when trained on our proposed normals. When visual inspecting the results shown in Figure~\ref{fig:normals_error} and Figure~\ref{fig:ladicky_error}, it is obvious that past papers have been training and evaluating on erroneous data. An easy way to see this is to check for color changes on planar surfaces such as walls.

\subsection{Improving Normals Using Point Clouds}
\label{sec:normals_pointclouds}

We propose to use the method introduced in \cite{holzer2012adaptive} to compute surface normals from a point cloud. We begin by smoothing the depth and filling holes as in \cite{Eigen_2015_ICCV} and then construct a 3D point cloud with PCL \cite{rusu20113d}. 
A key insight here is to use a large depth-adaptive smoothing window that adequately compensates for the noise introduced by the sensor, while not smoothing over visually salient depth discontinuities. While this is a straight-forward approach, it is important to note that it has not been done by any of the previous papers and our ablation studies show it greatly improves results. For this, we use the integral image approach implemented in PCL~\cite{holzer2012adaptive}. Compared to \cite{Silberman:ECCV12}, this samples a larger window more densely based off of the depth of the current point. We select a smoothing parameter of 30 for both real datasets (NYUDv2 and ScanNet) and 10 for the synthetic SceneNet dataset since it has minimal noise in its rendered depth estimates. 
We evaluate this method in Section~\ref{sec:eval}.

\subsection{Improving Normals with Semantics}
\label{sec:improve_normals}

To further reduce the amount of noise and get closer to absolute truth, we can leverage the semantics of the datasets. For certain semantic classes, e.g. walls and floors, we know that the results are usually planar (or at least piecewise planar). We use that information to smooth out the normals for those instances. Given that the datasets we are using all have some level of semantic information labeled, this is free contextual information.

While semantic segmentation gives us labels for objects, it does not distinguish between facets of the same class (for instance, walls facing in different directions). We perform an efficient post-processing step to identify regions with pixels that have consistent normals and semantic labels. We adapt the standard connected components algorithm to start at a pixel and grow the region outwards by adding pixels with normals that are within 30 degrees of the current averaged normal of the region, and of the same class as the starting pixel. We restrict this process to semantic labels that we assume to be planar. However, even if this assumption is violated, the normal variance constraint prevents arbitrary growth of these supposedly planar regions. Once we have computed the regions, we assign the averaged normal to all pixels of this region if the region is of a minimum size. An example of this is shown in Figure~\ref{fig:teaser}.
We evaluate this method in Section~\ref{sec:eval}.

\subsection{Evaluation}
\label{sec:eval}

\begin{table}[h]
\begin{center}
\small
\begin{tabular}{ |c|c|c|c|c|  }
\hline
 & \multicolumn{3}{|c|}{Accuracy}& \multicolumn{1}{|c|}{Error}\\
\hline
\textbf{Method} & $\leq 11.25 \degree$ &  $\leq 22.5 \degree$ & $\leq 30\degree$ & Mean Angle\\
\hline
Baseline & 46.2\% & 57.7\% & 63.8\% & 27.5\degree\\
Denoising & 49.5\% & 64.6\% & 71.1\% & 22\degree \\
Semantics & 60.6\% & 77.9\% & 83.4\% & 14.7\degree\\
\hline
\end{tabular}
\end{center}
\caption{
Accuracy and error rates of our normals model (without semantic output head)
when evaluated on the NYUDv2 normals from \cite{zeisl2014discriminatively}.
First row shows results using our simple network training on standard normals as shown in the top right of Figure \ref{fig:teaser}.
Second row shows results using our denoising method (bottom left of Figure \ref{fig:teaser}).
Third row shows results using our semantic smoothing method (bottom right of Figure \ref{fig:teaser}).
Evaluation is performed on the normals from \cite{zeisl2014discriminatively}, which demonstrates the necessity of a dataset cleanup.
\eat{
Ablations study on NYUDv2 by training on noisy normals from NYUDv2 (smoothing window of 10) vs. our proposed normals (window of 30) and finally the evaluation on the semantically corrected normals.
}
}
\label{table:different_gt} 
\end{table}

Table~\ref{table:different_gt} shows an ablation study done on NYUDv2 with different types of training data all evaluated on \cite{zeisl2014discriminatively} with the simple mobile encoder-decoder network described in Section~\ref{sec:details} without the performance increases we discuss later. The first row shows the results when trained on the normals used by many papers as shown in the top right of Figure \ref{fig:teaser}, the second row shows training on our normals computed from Section~\ref{sec:normals_pointclouds}, the final row shows training on our proposed ground truth surface normals that are semantically corrected. The normals from Section~\ref{sec:normals_pointclouds} result in much better accuracy compared to \cite{Eigen_2015_ICCV} or \cite{zeisl2014discriminatively}, especially in the larger angle errors and the mean angle error. Leveraging semantics improves results even more. This simple idea improves the mean angle error by almost 13 degrees and reduces the smallest angle errors by 14\%.

\begin{figure*}[htp]
	\includegraphics[width=\textwidth]{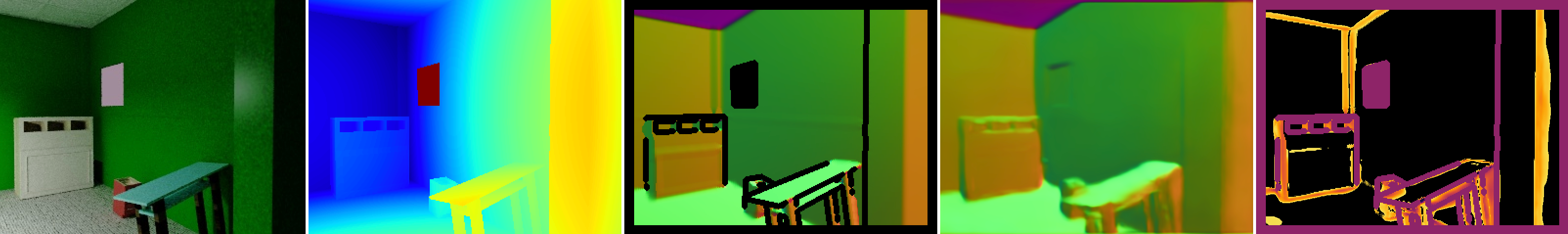}
\caption{
Qualitative evaluation on Scenenet data.
Columns, from left to right: RGB image, depth, ground truth surface normals, our predictions, error image (where black is under 11.25 degrees errors, and then error increases from yellow to purple).
}
\label{fig:scenenet}
\end{figure*}

\section{Combining synthetic and real data}
\label{sec:mixing_data}

We train and evaluate our network on several publicly available datasets, both real and synthetic, to reduce our dataset bias and produce more robust normals, as we explain below.

\subsection{Synthetic datasets}

\noindent\textbf{Scenenet}~\cite{mccormac2017scenenet} is a semi-photorealistic synthetic dataset of indoor environments comprised of $\sim 4$ million 320x240 images with corresponding depth and semantics. 
See Figure~\ref{fig:scenenet} for an example.
We compute the normals for Scenenet 
using the method proposed in Section \ref{sec:normal_computation};
however, we use a normal smoothing size of 10 given the input depth data is less noisy than data from conventional depth sensors.

\subsection{Mixing real and synthetic}

Since most prior work utilizes NYUDv2, our method was initially trained and evaluated only on it. However, we found this doesn't necessarily generalize well to other data,
as shown in the Supplementary.
(All results are obtained using the normals branch of the model architecture that is shown in
Figure~\ref{fig:architecture}; see Section~\ref{sec:details} for details.)

We also discovered that the standard practice of 
pretraining on Scenenet and finetuning on NYU
results in a model that generalizes poorly.
However, by simply mixing 10 synthetic scenenet images with 1 real image in every minibatch,
we were able to improve performance on both datasets.
Best results were obtained by mixing all 3 datasets, using
10 parts of Scenenet, 5 parts of Scannet, and 1 part of NYUDv2.
Qualitative results 
on Scenenet are shown in  Figure~\ref{fig:scenenet},
on NYU are shown in  Figure~\ref{fig:nyu},
and on Scannet in Figure~\ref{fig:scannet}.
See Supplementary for more examples.

\begin{table}[h]
\begin{center}
\small
\begin{tabular}{ |c|c|c|c|c|c|  }
\hline
 & \multicolumn{3}{|c|}{Accuracy}& \multicolumn{2}{|c|}{Error}\\
\hline
\textbf{Method} & 11.25 &  22.5 & 30 & Mean Angle & rmse\\
\hline
\cite{fouhey2013data}& 39.2 & 52.9 & 57.8 & 35.2 & -\\
\cite{zeisl2014discriminatively} & 27.7 & 49.0 & 58.7 & 33.5 & -\\
\cite{dharmasiri2017joint} & 44.9 & 67.7 & 76.3 & 20.6 & -\\
\cite{wang2016surge} & 47.3 & 68.9 & 76.6 & 20.6 & -\\
\cite{qi2018geonet} & 48.4 & 71.5& 79.5 & 19 & 26.9\\
\hline
Ours & 48.9 & \textbf{72.3} & \textbf{81.2} & \textbf{17} & 30.2\\
Ours on NYU' & \textbf{59.5} & \textbf{72.2} & 77.3 & 19.7 & \textbf{19.3}\\
\hline
\end{tabular}
\end{center}
\caption{Comparison against state of the art
on surface normal estimation task.
All methods (except in the last row) are evaluated on the normals from the NYUDv2 dataset
computed using the method of \cite{zeisl2014discriminatively};
in the last row, we show our method evaluated on the normals from NYU computed using our method,
which we denote by NYU'.
}
\label{table:sota} 
\end{table}

\eat{

During experimentation, we found that training on Scannet actually gives a mean angle error 1.6\% less on NYUDv2 compared to training only on NYUDv2. This bias can also be seen in the Scannet evaluation when trained only on NYUDv2. Mean angle error is 37.3 degrees, which is a large amount of error. That is comparable to the error evaluated on Scannet when trained on only Scenenet synthetic data (37.2 degrees). Results evaluated on Scenenet can be seen in Figure~\ref{fig:scenenet_normals}.
Finetuning on NYUDv2 from Scenenet synthetic data leads to similar dataset bias.

We discovered that mixing the synthetic data with real data and training from scratch gave us the best results. This can be seen in the Scenenet+NYU column in Table~\ref{table:datasets} and is shown on NYU in Figure~\ref{fig:nyu_normals}. We use 10 synthetic images for every 1 real image as there are 10x more synthetic images than real. Doing this improves the results on NYUDv2 by 2.6\% mean angle error and improves the percent of pixels with
more than 11.25\degree error by 8\%, a substantial improvement. This improves the results testing on synthetic data as well, though not as much. In order to generalize even more, we combine a training set of Scenenet (10 parts), Scannet (5 parts), and NYUDv2 (1 part), which achieves even better results on the real datasets. Results on Scannet are shown in Figure~\ref{fig:scannet_normals}.
}

\begin{table}[h]
\begin{center}
\small
\begin{tabular}{ |c|c|c| }
\hline
 & \multicolumn{2}{|c|}{Training set}\\
\hline
\textbf{Accuracy} & Scenenet+NYU FT & Datasets Mixed\\
\hline
\textbf{NYU} \% $<$ 11.25 & 57 & \textbf{59.5}\\
\% $<$ 22.5 & 69.6 & \textbf{72.2}\\
\% $<$ 30 & 74.6 & \textbf{77.3}\\
Mean Angle Error & 21.3 & \textbf{19.7}\\
\hline
\textbf{Scannet} \% $<$ 11.25 & 36.7 & \textbf{50.1}\\
\% $<$ 22.5 & 54.6 & \textbf{63.2}\\
\% $<$ 30 & 60.9 & \textbf{68.2}\\
Mean Angle Error & 34.1 & \textbf{28.8}\\
\hline
\textbf{Scenenet} \% $<$ 11.25 & 21 & \textbf{64.5}\\
\% $<$ 22.5 & 48.2 & \textbf{70.7}\\
\% $<$ 30 & 59.9 & \textbf{68.2}\\
Mean Angle Error & 37.6 & \textbf{26.1}\\
\hline
\end{tabular}
\end{center}
\caption{Normal Accuracy comparisons with different testing and training datasets. The columns are the training set used and the rows are the evaluation accuracy for each individual dataset. 
Scenenet+NYU FT means the standard practice of pretrained on synthetic and finetuned on NYUDv2. Datasets Mixed means the dataset is trained all from scratch with a batch-wise mix. The best result for each row is bold.}
\label{table:datasets} 
\end{table}

\subsection{Comparison with the state-of-the-art}

 In Table~\ref{table:sota}, 
 we show our results
 compared to previous state of the art surface normal estimation methods. Previous methods compute normals using
 the method of \cite{Silberman:ECCV12} or \cite{zeisl2014discriminatively}
 applied to various datasets.
 Here we compare all methods by evaluating on NYUDv2;
 for baseline methods, we 
 compute normals
 using the method of \cite{zeisl2014discriminatively},
 whereas for our method,
 we use our approach for computing normals during training.
 For testing, all methods 
 use the method of \cite{zeisl2014discriminatively} to compute normals.
 We outperform the previous state-of-the-art (SOTA), 
 \cite{qi2018geonet}, despite using a much smaller model,
due to the higher quality of our data.
For completeness, we also report the 
performance of our method
when evaluated on our proposed method of computing normals from NYU; this test set is more accurate, and is also more similar to training, so we see performance is even greater.

\begin{figure}[h]
	\includegraphics[width=0.5\textwidth]{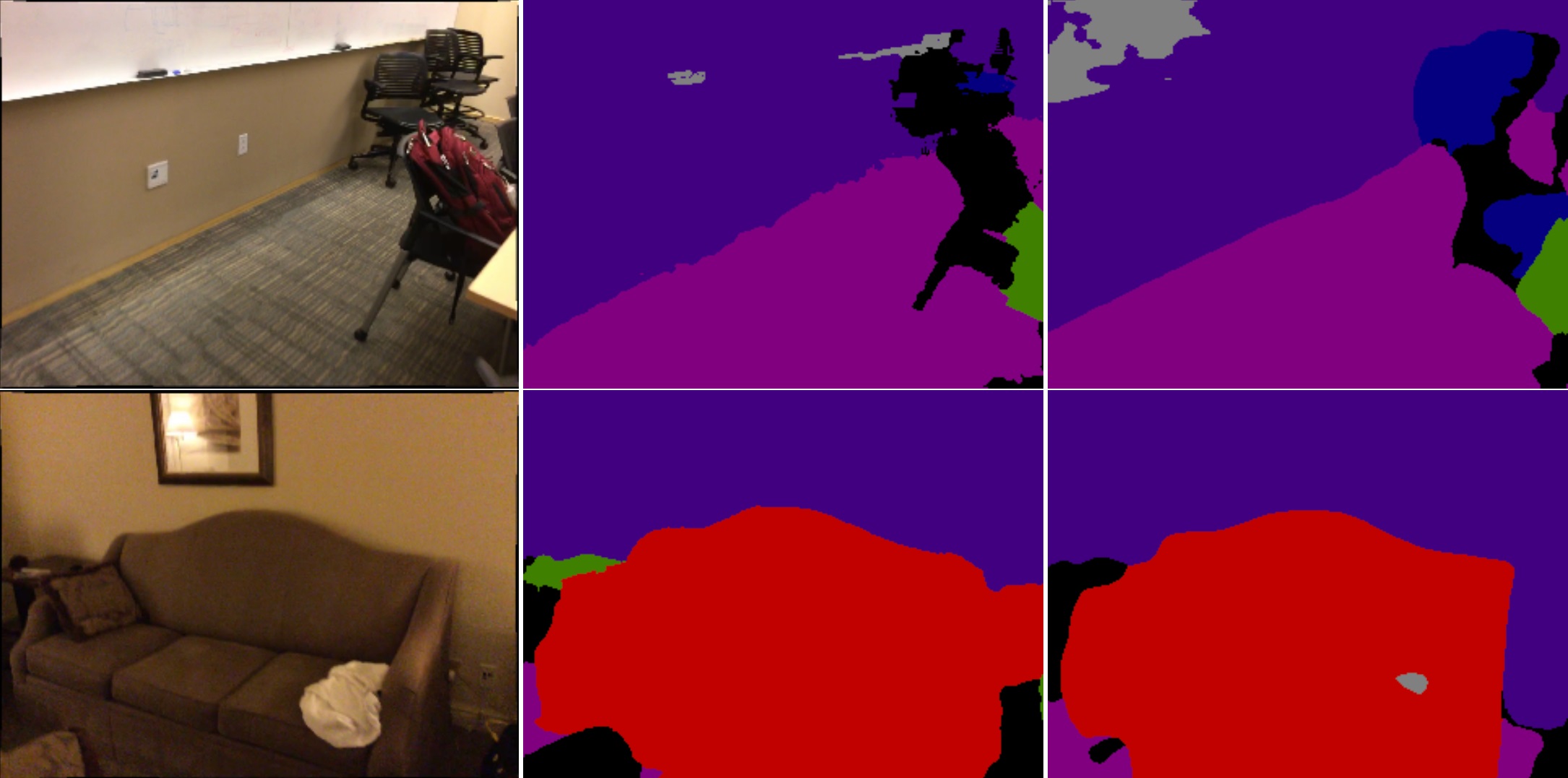}
\caption{Examples of our semantic labeling predictions for the Scannet dataset. From left to right: RGB image, ground truth, our predictions.}
\label{fig:scannet_semantics}
\end{figure}

\section{Jointly predicting semantics and normals}
\label{sec:joint}


In order to evaluate the effect of combining normals and semantic labeling, we did an ablation study using just the Scannet dataset. We chose it because it is a large, complex, real dataset with both normals and semantic labels. We used the 13 semantic labels from NYUDv2 for our experiments. These consist of bed, books, ceiling, chair, floor, furniture, objects, picture, sofa, table, tv, window, and wall. To fairly compare, we train on only Scannet in this ablation study, and ignore other datasets.

\begin{figure*}[t]
	\includegraphics[width=\textwidth]{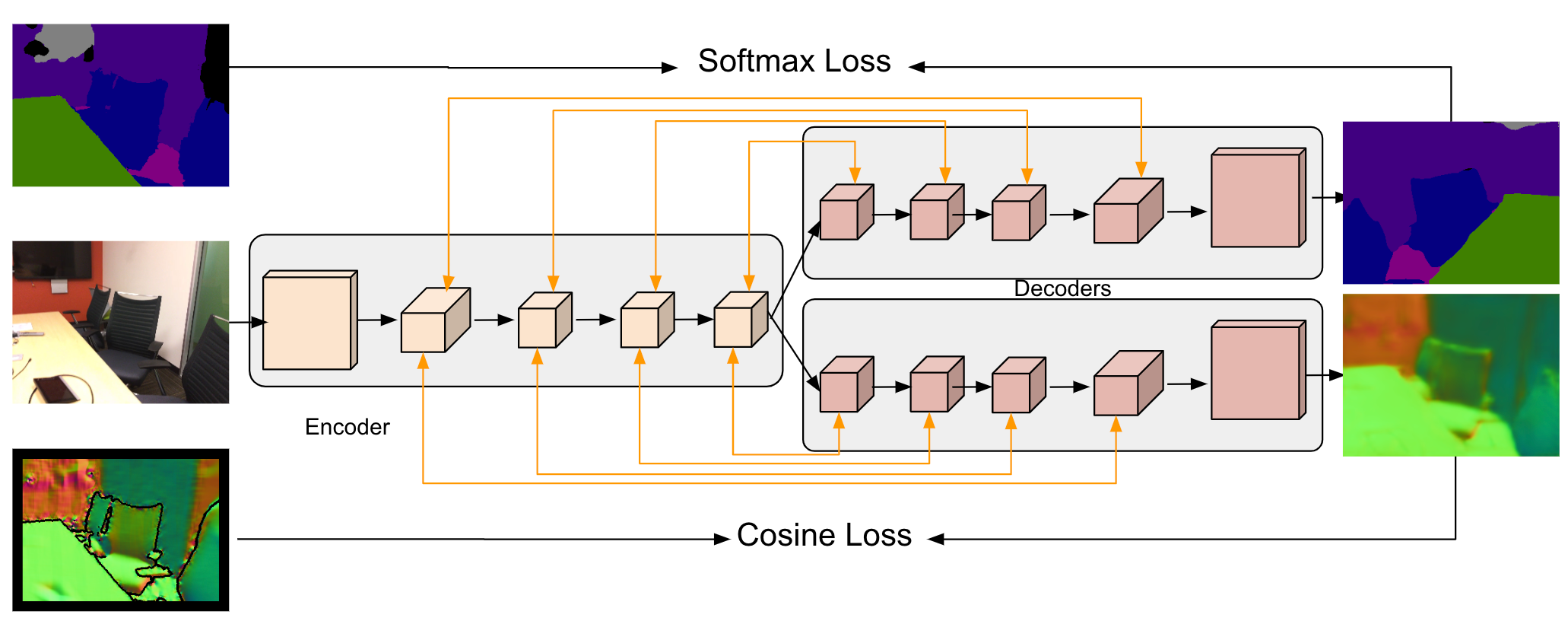}
\caption{Our Architecture for joint prediction involves a shared encoder inspired by Mobilenet~\cite{sandler2018mobilenetv2}, followed by two U-net decoders. Each outputs its prediction and has a separate loss for either segmentation or normal prediction. The losses along with regularization are summed and optimized jointly. When doing just normal prediction, we simply drop the segmentation decoder and loss. See Section~\ref{sec:details} for details.
}
\label{fig:architecture}
\end{figure*}

\subsection{Semantics}
In order to train semantic labeling, we use our same architecture and training with slight changes. We change the regression output of 3 channels with a cosine loss to a classification output of 14 labels using a softmax cross-entropy loss. Experimentally, semantic labeling seems to be a harder task than surface normal estimation, so in order to train, we finetune our whole architecture from our normals prediction network with a lower learning rate (0.001). Results showing the need for this are included in the Supplementary material. Semantic only prediction is shown in the Semantics column of Table~\ref{table:joint_prediction}.
See Figure~\ref{fig:scannet_semantics} for some qualitative results. Note that there is occasional error in the ground truth of the semantics as well. Even though semantics are just an intermediate task for our method, our results are still very promising. Our prediction in Figure~\ref{fig:scannet_semantics} correctly predicts both chairs as chairs (blue), even though the ground truth doesn't have this labeled correctly.

\subsection{Joint Prediction}
To train our method jointly, we duplicate the decoder using the architecture shown in Figure~\ref{fig:architecture}. We then finetune both the encoder and the dual decoders using the weights from our normal prediction network. The cosine and softmax cross-entropy losses are summed with a 20x weight modifier given to the cosine loss to balance them. An ablation study demonstrating why this is used is in the Supplementary material.

Contrary to prior work with joint decoders used for classification \cite{kokkinos2017ubernet}, our network improves when combining both semantics and normals. This is shown in Table~\ref{table:joint_prediction}. Surface normal estimation improves slightly (though it is important to note that small changes in the surface normal accuracy can still make large differences in practice due to the difference in angle error being so noticable when wrong). Interestingly, semantic labeling gets a large 6\% increase in pixel accuracy. We hypothesize this is due to the importance of shape as a cue for semantic labels. Note that this actually outperforms the previous results evaluated on Scannet in Table~\ref{table:datasets} as well. It's possible that without new normals proposed in Section ~\ref{sec:normal_computation}, this performance increase would not happen.

\begin{table}[h]
\begin{center}
\small
\begin{tabular}{ |c|c|c|c|  }
\hline
 & \multicolumn{3}{|c|}{Method}\\
\hline
\textbf{Accuracy} & Normals & Semantics & Joint\\
\hline
\% $<$ 11.25 & 49.3 & N/A & \textbf{50.9}\\
\% $<$ 22.5 & 63.2 & N/A & \textbf{65.2}\\
\% $<$ 30 & 68.2 & N/A & \textbf{70}\\
Mean Angle Error & 29 & N/A & \textbf{28}\\
Semantic Accuracy \% & N/A & 59 & \textbf{65.6}\\
\hline
\end{tabular}
\end{center}
\caption{This shows the results of Joint semantics and normals prediction on Scannet. The Normals column is the accuracy of a network only trained on Scannet normals. The semantics is the accuracy when only trained on Scannet semantics. Joint is the accuracy when both are trained concurrently as per Section~\ref{sec:joint}.}
\label{table:joint_prediction} 
\end{table}

\vspace*{-0.7em}
\section{Training a realtime model}
\label{sec:details}

In this section, we describe how to use the above techniques, combined with a lightweight model, to build a real-time mobile system with state of the art accuracy. We discuss our model size, training pipeline, and small tricks (reducing the number of channels and utilizing grayscale) used to get the model on a mobile device.

\subsection{Model}\label{sec:architecture}
Prior approaches to the task of normal prediction have used feature extractors trained on VGG~\cite{simonyan2014very} or ResNet~\cite{he2016deep}. In contrast, we use a light-weight architecture that lends itself well to mobile applications. For our surface normal experiments and ablation studies, we use a modified version MobileNetV2 ~\cite{sandler2018mobilenetv2} encoder followed by the U-net decoder~\cite{ronneberger2015u}. The key changes to MobileNetV2 are a normal residual instead of the inverted residual, PReLU~\cite{he2015delving} instead of ReLU, removing global average pooling, and increasing the convolutional filter size to 5.

For the decoder, we use U-net with 4 bilinear resizes, convolutions, and concatenations (see Supplementary material). These correspond to the blocks in MobileNetV2. The final output of the decoder is resized to the  width and height of the input image (320 x 240 in our experiments), with the number of channels defined by the output task (i.e. 3 for normals and 14 for semantic labeling of NYU13).

 After training our network, we can remove unnecessary ops and only use the normals encoder-decoder path by converting that model to a flatbuffer using Tensorflow Lite\cite{tflite}. Our final network is under 2MB in size. We run inference using this model on the phone via ops implemented as OpenGL shaders. 

\subsection{Finetuning vs. Training from scratch}

Conventionally, encoder-decoder networks use a larger encoder like ResNet101 (which is pretrained on Imagenet) and then fine-tune them for the specific task. However, for the task of surface normal estimation, we found that training from scratch in an end-to-end manner gave us better results. This could be due to the Imagenet dataset bias, our small network encoder, or the uniqueness of the task. 


For training, when learning only surface normals in a single architecture as in our ablation studies, we use RMSProp~\cite{tieleman2012lecture} with a weight decay of 0.98, a learning rate of 0.045. When we train on surface normals and semantics, we fine tune off the surface normals model with a lower learning rate of 0.001. 

\begin{table}[h]
\begin{center}
\small
\begin{tabular}{ |c|c|c|c|  }
\hline
 & \multicolumn{3}{|c|}{\% of images grayscale}\\
\hline
\textbf{Accuracy} & 0\% & 50\% & 100\%\\
\hline
\textbf{NYU} \% $<$ 11.25 & 59.1 & 59.2 & 59.1\\
\% $<$ 22.5 & 72.2 & 72.2 & 72.2\\
\% $<$ 30 & 77.4 & 77.3 & 77.3\\
Mean Angle Error & 19.5 & 19.4 & 19.5\\
\hline
\textbf{Scannet} \% $<$ 11.25 & 49.6 & 49.5 & 49.6\\
\% $<$ 22.5 & 63.6 & 63.6 & 63.4\\
\% $<$ 30 & 68.6 & 68.6 & 68.5\\
Mean Angle Error & 28.8 & 28.8 & 28.9\\
\hline
\textbf{Scenenet} \% $<$ 11.25 & 60.7 & 63.2 & 63.3\\
\% $<$ 22.5 & 70.3 & 70.6 & 70.5\\
\% $<$ 30 & 72.9 & 72.9 & 72.8\\
Mean Angle Error & 27.1 & 26.5 & 26.5\\
\hline
\textbf{Average} \% $<$ 11.25 & 57.075 & 57.625 & 57.7\\
\% $<$ 22.5 & 69.025 & 69.05 & 68.975\\
\% $<$ 30 & 73.325 & 73.3 & 73.225\\
Mean Angle Error & 24.225 & 24.075 & 24.125\\
\hline
\end{tabular}
\end{center}
\caption{This ablation study shows the effect of color on the network by changing the percent of input training images that are converted to grayscale. Interestingly enough, color does not seem that important for surface normal estimation.}
\label{table:grayscale} 
\end{table}

\subsection{RGB vs. Grayscale}
Interestingly enough, for the task of surface normal estimation, color doesn't give much more of an advantage over grayscale data. This is shown in Table~\ref{table:grayscale}. This suggests that the neural network learns edges and luminance rather than specific colored features. This can potentially reduce the size and number of operations in a network. An ablation study on this is shown in Table ~\ref{table:grayscale}.

\begin{table}[h]
\begin{center}
\small
\begin{tabular}{ |c|c|c|c|c|  }
\hline
 & \multicolumn{4}{|c|}{Channel Multiplier}\\
\hline
\textbf{Accuracy} & 12 & 16 & 22 & 32\\
\hline
\textbf{NYU} \% $<$ 11.25 & 56.1 & 57.1 & 58 & 59.3\\
\% $<$ 22.5 & 67.7 & 68.6 & 69.3 & 69.6\\
\% $<$ 30 & 72.3 & 73.1 & 73.7 & 73.9\\
Mean Angle Error & 22.8 & 22.3 & 22 & 21.8\\
\hline
\textbf{ScanNet} \% $<$ 11.25 & 44.5 & 46 & 47.6 & 50.1\\
\% $<$ 22.5 & 60.3 & 61.4 & 62.3 & 63.2\\
\% $<$ 30 & 65.9 & 66.9 & 67.6 & 68.2\\
Mean Angle Error & 30.6 & 30 & 29.5 & 28.8\\
\hline
\textbf{SceneNet} \% $<$ 11.25 & 59.9 & 61.5 & 62.3 & 64.5\\
\% $<$ 22.5 & 68.8 & 69.6 & 70 & 70.7\\
\% $<$ 30 & 71.4 & 72.1 & 72.4 & 78.2\\
Mean Angle Error & 27.8 & 27.2 & 26.9 & 26.1\\
\hline
\textbf{Average Eval} \% $<$ 11.25 & 54.08 & 55.13 & 56.48 & 58.15\\
\% $<$ 22.5 & 66.55 & 67.35 & 68.20 & 68.63\\
\% $<$ 30 & 71.10 & 71.85 & 72.53 & 74.10\\
Mean Angle Error & 25.55 & 25.05 & 24.63 & 24.20\\
\hline
\textbf{\# million MpAdds} & 467 & 673 & 987 & 1624\\
\hline
\end{tabular}
\end{center}
\caption{Here we use an ablation studty to test performance vs network size. The channel multiplier is a multiplier that determines the number of output channels calculated at each block. For instance, the final output of the encoder at channel multiplier 32 has 1280 channels, whereas, at channel multiplier 16, it has 640 channels.}
\label{table:channel_multiplier} 
\end{table}

\subsection{Network Size}
In Mobilenet~\cite{sandler2018mobilenetv2}, an encoder-decoder architecture is proposed with network size and speed described in the number of multiply-adds (MpAdds). We also test our normal prediction network as a function of network size in the same manner. 
The results are shown in
Table~\ref{table:channel_multiplier}.

A semantic segmentation model was also proposed by ~\cite{sandler2018mobilenetv2} with DeepLab as a decoder, where the last encoder layer is removed to minimize model size. This model is 2.75B MpAdds with stride 16 and 152.6B MpAdds with stride 8. We found that it is actually better to keep the last layer and just reduce the channel size of each layer; this results in a faster and smaller model. Our proposed network has 1.624B MpAdds at its largest, and only 467M for the mobile version, 
which is almost 6x smaller than the stride 16 version and more than 300x smaller than the stride 8 version. 
The other SOTA methods we compare against earlier in the paper, that utilize Resnet-101 or VGG-19, have between 91B and 5000B MpAdds, which is several orders of magnitude larger than our proposed network even though our method has better results.


\subsection{Applications}

\begin{figure}[h]
\centering
\includegraphics[width=110px]{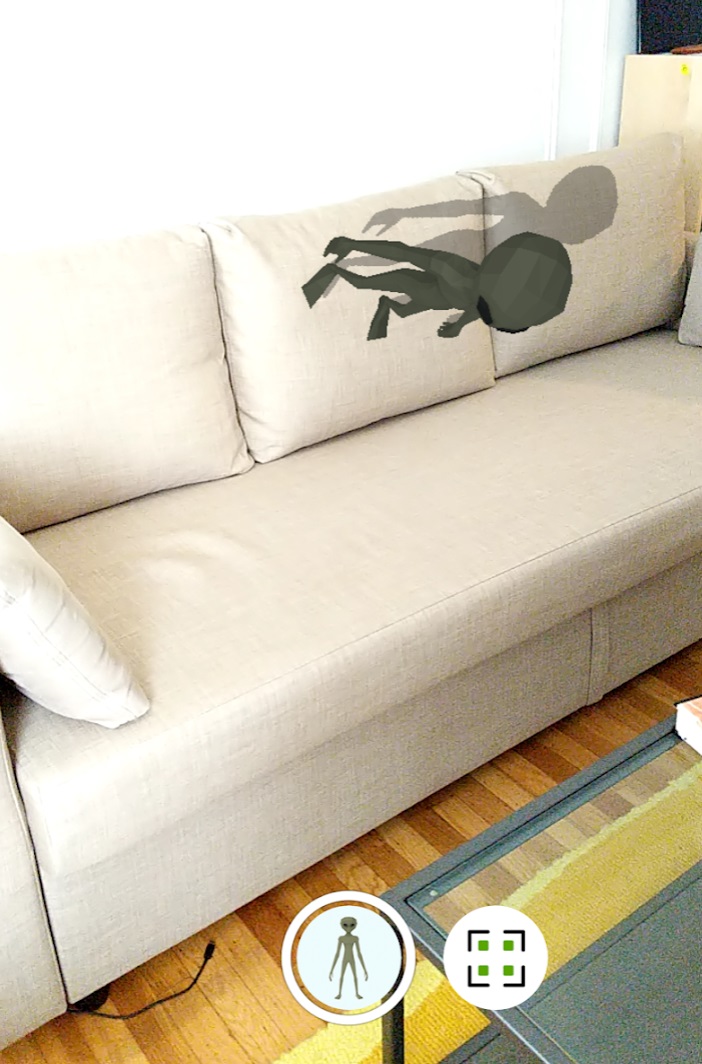}
\includegraphics[width=110px]{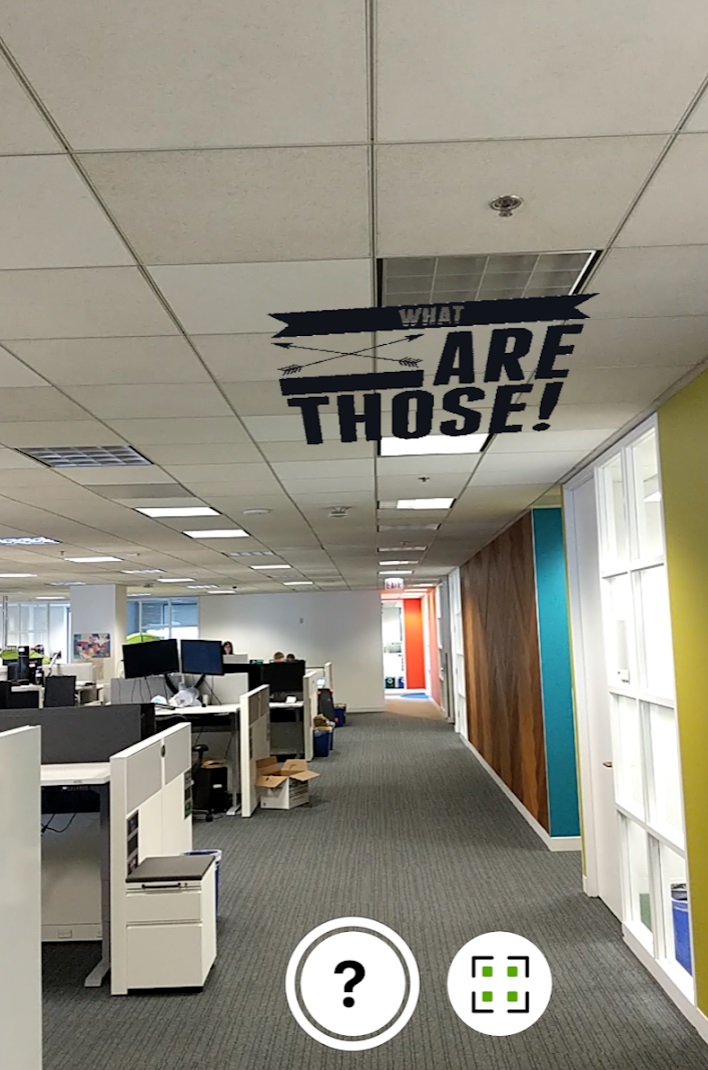}
\caption{A sample AR application that uses the surface orientation to place a virtual character and text.}
\label{fig:alien}
\end{figure}

Using the channel scaling and other model minimization techniques discussed above, we created a lightweight architecture that can be run at 12fps on a mobile phone. To demonstrate our state of the art results on normal estimation in real-time, we use this estimation to place stickers on surfaces in their natural orientation. A video of this demo is included in the supplementary material. Screenshots showing this demo running on the mobile device are shown in Figure ~\ref{fig:alien}. By simply clicking a region, the sticker or object can be placed realistically in AR using the predicted surface normals.

\section{Conclusion}

We have shown several simple methods for significantly improving the accuracy of any CNN method for predicted surface normals, namely:
calculate the ground truth normals in a better way;
combine real and synthetic data in a better way;
and jointly train for normal prediction and semantic segmentation.
We have also shown how to use these ideas to train a lightweight model that gives state of the art results,
has low memory footprint, and runs at interactive rates on a mobile phone. In addition to this, we plan on releasing our new computed ground truth normals.

\eat{
We show that correcting ground truth surface normals drastically improves the performance
of CNNs trained for surface normal estimation. We also show that properly using synthetic data improves these results. Furthermore,
by using semantic labels, 
not only can we improve the normals but we can train a network to jointly predict both normals and semantics. This method provides increased accuracy for both tasks. In addition to this, our normal estimation method has a low memory footprint and runs at interactive rates on a mobile phone.
}

{\small
\bibliographystyle{ieee}
\bibliography{egbib}
}

\end{document}